\documentclass[letterpaper,10pt,conference]{ieeeconf}
\IEEEoverridecommandlockouts 
\usepackage{url}
\usepackage{graphicx}
\usepackage{cite}
\makeatletter
\let\NAT@parse\undefined
\makeatother
\usepackage{amsmath}
\usepackage{amssymb}  
\usepackage{hyperref}
 \usepackage[aboveskip=1pt]{subcaption}
\usepackage[skip=0.5pt]{caption}
\usepackage{multirow}
\usepackage{multicol}
\usepackage{booktabs}
\usepackage{color}
\usepackage{xcolor}
\usepackage[ruled,vlined,noend]{algorithm2e}
\providecommand{\keywords}[1]{\textbf{\textit{Index terms---}} #1}
\begin{document}
%
\title{\bf{Deep Kernel and Image Quality Estimators for Optimizing Robotic Ultrasound Controller using Bayesian Optimization}}
\author{Deepak Raina$^{12*}$, SH Chandrashekhara$^3$, Richard Voyles$^2$, Juan Wachs$^2$, Subir Kumar Saha$^1$\\\vspace{-0.5cm}
\thanks{This work was supported in part by SERB (India) - OVDF Award No. SB/S9/Z-03/2017-VIII; PMRF - IIT Delhi under Ref. F.No.35-5/2017-TS.I:PMRF; National Science Foundation (NSF) USA under Grant \#2140612; Daniel C. Lewis Professorship and PU-IUPUI Seed Grant.}
\thanks{$^{1}$Indian Institute of Technology (IIT), Delhi, India (\{deepak.raina, saha\}@mech.iitd.ac.in); $^{2}$Purdue University (PU), Indiana, USA (\{draina, rvoyles, jpwachs\}@purdue.edu); $^{3}$All India Institute of Medical Sciences (AIIMS), Delhi, India (drchandruradioaiims@gmail.com).}
}
\maketitle
\begin{abstract}
Ultrasound is a commonly used medical imaging modality that requires expert sonographers to manually maneuver the ultrasound probe based on the acquired image. Autonomous Robotic Ultrasound (A-RUS) is an appealing alternative to this manual procedure in order to reduce sonographers' workload. The key challenge to A-RUS is optimizing the ultrasound image quality for the region of interest across different patients. This requires knowledge of anatomy, recognition of error sources and precise probe position, orientation and pressure. Sample efficiency is important while optimizing these parameters associated with the robotized probe controller. Bayesian Optimization (BO), a sample-efficient optimization framework, has recently been applied to optimize the 2D motion of the probe. Nevertheless, further improvements are needed to improve the sample efficiency for high-dimensional control of the probe. We aim to overcome this problem by using a neural network to learn a low-dimensional kernel in BO, termed as Deep Kernel (DK). The neural network of DK is trained using probe and image data acquired during the procedure. The two image quality estimators are proposed that use a deep convolution neural network and provide real-time feedback to the BO. We validated our framework using these two feedback functions on three urinary bladder phantoms. We obtained over 50\% increase in sample efficiency for 6D control of the robotized probe. Furthermore, our results indicate that this performance enhancement in BO is independent of the specific training dataset, demonstrating inter-patient adaptability.  
\end{abstract}

%
\IEEEpeerreviewmaketitle

\section{Introduction}
Ultrasound imaging delivers instantaneous, non-invasive access to the human body to effectively visualize anatomy and pathology \cite{chan2011basics}. Ultrasound stands out among other imaging modalities, such as MRI or CT scan, because it is low-cost, non-radiating, and suitable for use in rural or underserved communities. However, unlike other modalities, ultrasound is not a plug-and-play technology. Acquiring quality images relies heavily on the expertise of the trained sonographer who ensures optimal ultrasound probe contact location, pressure, angle, and gel condition between the probe and skin \cite{raina2021comprehensive}. In addition, the sonographer must integrate the visual feedback from the image with a mental model of anatomical structure(s) to continuously improve the probing process. 



\begin{figure}[t]
	\centering
	\includegraphics[trim=0cm 5cm 8cm 0cm,clip,width=\linewidth]{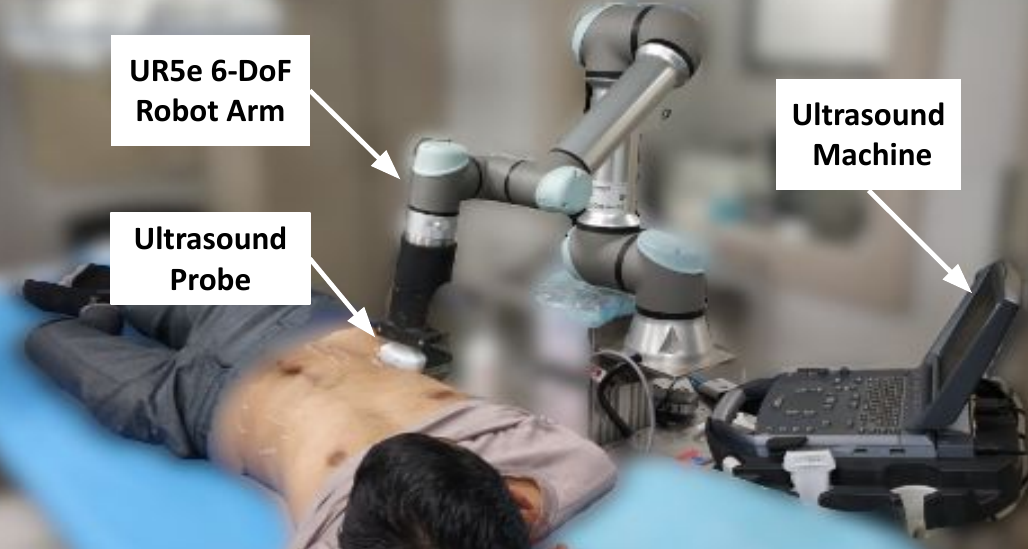}
	\caption{Robotic ultrasound system with probe attached to its end-effector \cite{raina2021comprehensive}, conducting an \textit{in-vivo} ultrasound.}
	\label{fig:real_exp_setup}
\end{figure}

In the pursuit of reducing the dependency on experts, enhancing access to care and eliminating the requirement for direct patient contact, a Robotic Ultrasound System (RUS) is introduced \cite{roshan2022robotic}. RUS has a 6-DoF robotic arm with an ultrasound probe attached to its end-effector, as shown in Fig. \ref{fig:real_exp_setup}. Recently, several telerobotic or human-assisted ultrasound systems have been proposed \cite{raina2021comprehensive, duan20215g, carriere2019admittance, li2022dual}.
In contrast to those systems, our system is fully autonomous. This allows for shorter procedural times, lower cognitive load, and prevents communication delays. However, several challenges need to be addressed for autonomous RUS. One of the key challenges is high-dimensional control of the probe maneuvers over the patient body. The probe position, angle and pressure need to be adjusted precisely and uniquely for inter- and intra-patient procedures. Inappropriate probe maneuvers would lead to acquiring low-quality images having artifacts, thereby leading to false alarms during the patients' diagnosis \cite{pinto2013sources}. In order to master this procedure's skill, novice operators repeatedly scan different subjects in laboratory and clinical setting during their medical training \cite{tarique2018ultrasound}. Inspired by their training method, several model-based \cite{chatelain2017confidence, jiang2020automatic} and model-free \cite{li2021autonomous, li2021image,ning2021autonomic} paradigms have been applied to learn the robotized probe maneuvering for acquiring the optimal image. However, these techniques either provide limited probe motions or require a prohibitive number of samples for training, making them infeasible for in-human optimization of the RUS controller. This has led to the research community's interest in sample-efficient learning techniques for optimizing the RUS controllers.

Bayesian Optimization (BO) is a popular sample-efficient, gradient-free and black-box optimization method for learning robotic controller parameters \cite{roveda2021human, zahedi2022user}. Recently, BO has been applied to optimize the 2D motion of robotized ultrasound probe \cite{goel2022autonomous}. The clinical procedures, however, require high-dimensional motion of probe for adjusting its position, orientation and forces.  Moreover, the performance of BO degrades in high dimensions due to its exponentially increasing statistical and computational complexity \cite{choffin2018scaling, malu2021bayesian}. 
In this work, we formulate a transformation that uses a neural network to construct a low-dimensional kernel metric, thereby providing a sample-efficient BO framework for optimizing a high-dimensional RUS controller. The kernel metric is termed as Deep Kernel (DK), which is learned by utilizing the probe and image data obtained offline from the ultrasound procedure. This requires maneuvering the ultrasound probe over the scanning region and recording the probe poses and corresponding images. The neural network then learns a mapping between the high-dimensional probe pose and low-dimensional image quality metric. In order to estimate image quality, we propose two estimators using a Deep Convolution Neural Network (D-CNN) for image classification and segmentation, respectively, which also provide real-time feedback to BO. In order to validate it experimentally, we optimized a 6D control of the robotized probe, including position, orientation and force along z-axis, for three urinary bladder phantoms. We demonstrate that using a Deep Kernel in BO outperforms the sample-efficiency of standard kernel formulation. We noted that DK's performance is independent of a specific training dataset, which shows inter-patient adaptability. Finally, we validated the proposed image quality estimators on the phantom dataset and noted enhancement in accuracy over state-of-the-art networks.
\subsection{Related Work}
\subsubsection{Autonomous Robotic Ultrasound (A-RUS) systems}
Several A-RUS systems have been developed to address the challenges of manual procedure. Recent works have used confidence-maps \cite{karamalis2012ultrasound, chatelain2017confidence, jiang2020automatic} or support vector machines \cite{akbari2021robotic} to assess image quality and then used it for adjusting the probe poses and forces. However, the probe control is limited, having either $2$D/$3$D motion or decoupled poses/forces, which is insufficient to acquire a good quality ultrasound image for complex procedures on patients with challenging physiological conditions.  
Li \textit{et. al.} \cite{li2021autonomous, li2021image} have used a deep reinforcement learning framework to develop a robotized probe controller for optimizing the image quality in spinal ultrasound. However, these systems require a large number of explorations to find the global optimum, therefore their success is limited to phantoms. Jiang \textit{et. al.} \cite{jiang2021autonomous, jiang2022towards} have used segmentation of tubular structures in human limbs for scanning them while the limb is static or moving. However, the procedures like spine or limb ultrasound require the probe to be oriented in a normal direction to the point of contact, with minimal orientation adjustment required throughout scanning. In contrast to these systems, we propose a generic A-RUS with a comprehensive probe controller using a sample-efficient BO framework.
\subsubsection{Bayesian optimization for ultrasound robots}
BO has been used for several safety-critical robotic medical procedures, such as autonomous robotic palpation \cite{yan2021fast}, semi-autonomous surgical robot \cite{chen2020supervised}, controller tuning of hip exoskeletons \cite{ding2018human} and A-RUS \cite{huang2021towards, goel2022autonomous}.
Our work is a non-trivial extension to work by Goel \textit{et al.} \cite{goel2022autonomous}. They proposed using BO for A-RUS utilizing segmentation of the vessel in the ultrasound image as feedback to the BO for finding and scanning the region with high vessel density. They used hybrid position-force control to move the robot in $2$D plane while maintaining constant force along the probe axis normal to the point of contact. In contrast, our work proposes technical enhancements to make this approach more practical and expand its scope to complex ultrasound procedures. We propose a $6$D control of the probe pose, including the variable force control along the $z-$axis. We further propose a novel kernel for BO to improve its sample efficiency during the optimization of $6$D probe controller. 
\subsubsection{High-Dimensional Bayesian Optimization}
A large body of literature has addressed the issues related to the scalability of BO to high-dimensional optimization problems. A few methods have exploited the potential additive structure in the objective function \cite{kandasamy2015high, wang2018batched}. However, these methods require partitioning of the objective function and a large number of GPs, hence they do not generalize well to several applications. The other methods determined the mapping from high-dimensional space to low-dimensional subspace \cite{wang2016bayesian, nayebi2019framework}. For robotic systems, the transformation of search space to a lower dimension has been attempted for locomotion controller \cite{antonova2017deep, rai2018bayesian}. Rai \textit{et. al.} \cite{rai2018bayesian} proposed a transformation based on human walking knowledge to project a $16$D controller to $1$D controller. Antonova \textit{et. al.} \cite{antonova2017deep} propose to learn the transformation using a neural network trained on data obtained from a high-fidelity simulator. However, the paradigms of high-dimensional BO have not yet been explored for optimizing the RUS controller.
\begin{figure*}[!ht]
	\centering
	\includegraphics[trim=0cm 10cm 0cm 0cm,clip,width=\linewidth]{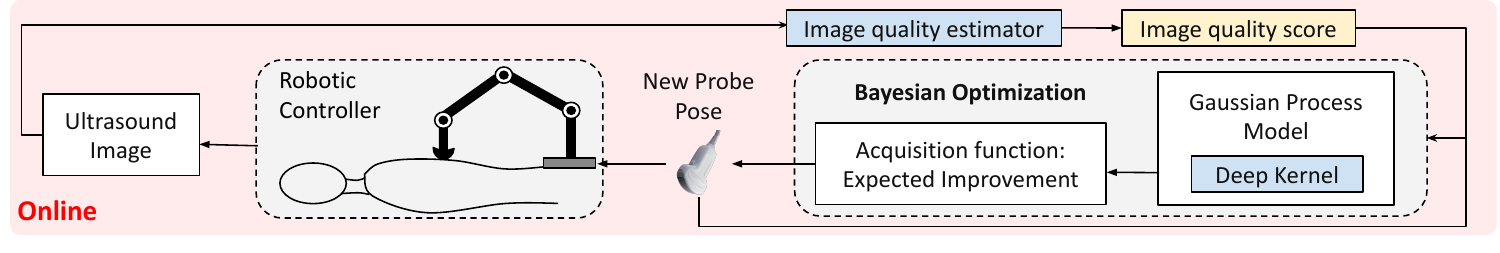}
	\caption{Overview of the Bayesian optimization (BO) framework for optimizing 6D robotic ultrasound controller using deep kernels in Gaussian process model and image quality estimators as feedback.}
	\label{fig:overview}
\end{figure*}
\section{Methodology} \label{sec:methodology}
The overview of the proposed BO is shown in Fig. \ref{fig:overview}. In the \textit{offline phase}, first the dataset of images will be collected from the scanning region by varying the probe poses. Then, each image in dataset will be annotated for quality rating and segmentation masks, which will be used to train the \textit{Image Quality Estimators (IQEs)}. Second, the \textit{Deep Kernel (DK)} neural network is trained using the dataset of probe poses and their corresponding image qualities estimated using the IQEs trained earlier. In the \textit{online phase}, the BO will utilize the DK in Gaussian Process (GP) model and learnt IQEs to calculate the probabilistic estimate of the unknown image quality in the scanning region of human body. An \textit{acquisition function} is optimized during online BO to yield the new probing pose, which will acquire the new ultrasound image. The GP model is then re-fitted to the new data and process is repeated until the termination criteria is reached, which can either be the maximum quality reached or maximum reasonable steps.
\subsection{Bayesian optimization}
The BO can determine the probing poses that will optimize the ultrasound image quality within a scanning region. If $A$ denotes the scanning region on the human body, then the objective function of BO is given by:
\begin{equation}
    \max_{\boldsymbol{p} \in A} q(\boldsymbol{\mathcal{I}}(\boldsymbol{p}))
\end{equation}
where $q(\boldsymbol{\mathcal{I}}(\boldsymbol{p}))$ denotes the ultrasound image quality score  $\boldsymbol{\mathcal{I}}$ at probing pose $\boldsymbol{p}$. 
\\
\vspace{-2mm}
\subsubsection{Deep Kernel} \label{sec:deep_kernel} 
The GP model is used as the function estimator in BO. It defines an unknown quality function value $q(\boldsymbol{\mathcal{I}}(\boldsymbol{p}))$ at a probe pose $\boldsymbol{p}$ by a mean function $\boldsymbol{\mu}(\cdot)$ and kernel function $\boldsymbol{\kappa}(\cdot,\cdot)$ as
\begin{equation}
    q(\boldsymbol{\mathcal{I}}(\boldsymbol{p})) \sim GP(\boldsymbol{\mu}(\cdot),\boldsymbol{\kappa}(\cdot,\cdot))
\end{equation}
Given the quality values set $\boldsymbol{\bar{q}} = [{q}(\boldsymbol{\mathcal{I}}(\boldsymbol{p}_1)), \cdots, {q}(\boldsymbol{\mathcal{I}}(\boldsymbol{p}_n))]$ at probe poses $\boldsymbol{\bar{p}} = [\boldsymbol{p}_1, \cdots, \boldsymbol{p}_n]$, the GP regression can predict the quality function value at new probe pose $\boldsymbol{p}^*$ as the Gaussian distribution and is given by:
\begin{equation}
    \mathcal{P}(q(\boldsymbol{\mathcal{I}}(\boldsymbol{p}^*)|\boldsymbol{p}^*,\bar{\boldsymbol{p}}, \bar{\boldsymbol{q}}) = \mathcal{N}(\boldsymbol{k} \boldsymbol{K}^{-1} \bar{\boldsymbol{q}}, \boldsymbol{\kappa}(\boldsymbol{p}^*, \boldsymbol{p}^*) - \boldsymbol{k} \boldsymbol{K}^{-1} \boldsymbol{k}^T)
\end{equation}
where,
\begin{equation*}
\boldsymbol{k} = \begin{bmatrix}
\boldsymbol{\kappa}(\boldsymbol{p}_*, \boldsymbol{p}_1) & \cdots & \boldsymbol{\kappa}(\boldsymbol{p}_*, \boldsymbol{p}_n)
\end{bmatrix} 
\end{equation*}
\begin{equation*}
    \boldsymbol{K} = \begin{bmatrix}
\boldsymbol{\kappa}(\boldsymbol{p}_1, \boldsymbol{p}_1) & \cdots & \boldsymbol{\kappa}(\boldsymbol{p}_1, \boldsymbol{p}_n)\\
\vdots & \ddots & \vdots \\
\boldsymbol{\kappa}(\boldsymbol{p}_n, \boldsymbol{p}_1) & \cdots & \boldsymbol{\kappa}(\boldsymbol{p}_n, \boldsymbol{p}_n)
\end{bmatrix}
\end{equation*}
For kernel matrix $\boldsymbol{\kappa}$, we propose using a sum of two functions, the radial basis function and white noise function, because their combination provides better estimates for anatomical structures in ultrasound images \cite{goel2022autonomous}.
\begin{equation} \label{eq:kernel}
   {\kappa}(\boldsymbol{p}_i, \boldsymbol{p}_j) = \sigma_r \exp \bigg(\frac{-||\boldsymbol{p}_i - \boldsymbol{p}_j||^2}{2l^2}\bigg) + \sigma_w \boldsymbol{{I}_n}
\end{equation}
where $\sigma_r$, $\sigma_w$ and $l$ is the overall variance, noise variance and length-scale, respectively, representing the BO hyperparameters $\boldsymbol{\theta}$. Robotized control of ultrasound probe necessitates high-dimensional control of probe pose, including its position, orientation and forces to acquire appropriate quality images for inter- and intra-patient procedures. However, the performance of BO often degrades in high dimensions \cite{malu2021bayesian}, specifically due to two reasons: (i) the high dimensional objective function exponentially increases the number of samples (queries) required to explore the space (ii) the BO's acquisition function is non-convex and optimizing it demands exponentially increasing computational power. 
\begin{figure}[!ht]
	\centering
	\includegraphics[trim=0cm 8cm 16cm 0cm,clip,width=0.8\linewidth]{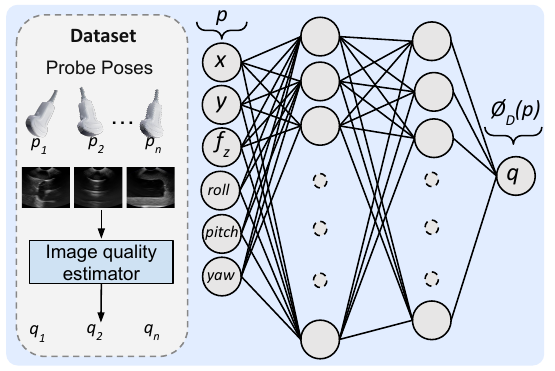}
	\caption{Dataset and neural network structure for Deep Kernel}
	\label{fig:deep_kernel}
\end{figure}

In order to overcome these issues, we propose a $1$D kernel for GP and term it as a Deep Kernel, as shown in Fig. \ref{fig:deep_kernel}. It will transform the 6D representation of probe poses into a 1D space. We use a supervised Neural Network (NN), represented by $\boldsymbol{\phi}_D$, which will learn the mapping between probe pose representation and image quality metric. The DK neural network used two hidden layers, each having $400$ and $300$ neurons, respectively. The activation function used for each layer is Rectified Linear Unit (ReLU). The estimation of image quality metric for a given image has been explained in section \ref{sec:quality_estimator}. The data for training the neural network can be collected offline by the expert-demonstrated probing maneuvers or random sampling of the probe poses over the scanning region.
The functional form of this kernel is identical to that of eq. \eqref{eq:kernel}. However, the euclidean difference between the six-dimensional probe poses was replaced with the difference between the uni-dimensional NN output image quality score for the input probe pose.
\begin{equation} \label{eq:deep_kernel}
   {\kappa}(\boldsymbol{p}_i, \boldsymbol{p}_j) = \sigma_r \exp \bigg(\frac{-||\boldsymbol{\phi}_D(\boldsymbol{p}_i) - \boldsymbol{\phi}_D(\boldsymbol{p}_j)||^2}{2l^2}\bigg) + \sigma_w \boldsymbol{{I}}
\end{equation}
where, $\boldsymbol{\phi}_D(\boldsymbol{p}_*)$ denotes the output of NN at probe pose $\boldsymbol{p}_*$.
The hyper-parameters $\boldsymbol{\theta}$ of DK-based GP will be estimated by maximizing the log marginal likelihood with a Limited memory Broyden–Fletcher–Goldfarb–Shanno (L-BFGS) solver, which is given by
\begin{equation}
    \boldsymbol{\theta}^* = \arg\max_{\boldsymbol{\theta}} \log \prod \mathcal{N} (\boldsymbol{q}(\mathcal{I}(\boldsymbol{p}_i))| \boldsymbol{\mu}_{\boldsymbol{\theta}}(\boldsymbol{p}_i), \boldsymbol{K})
\end{equation}
\subsubsection{Acquisition Function} The next probe pose is determined by optimizing the acquisition function. We have used an Expected Improvement (EI), which is the most commonly used acquisition function. Given the posterior mean and variance of GP as $\boldsymbol{\mu}_{\bar{\boldsymbol{q}}}(\boldsymbol{p}), \boldsymbol{\sigma}_{\bar{\boldsymbol{q}}}^2(p)$, the expression for EI is:
\begin{equation} \label{eq:ei}
    \resizebox{0.9\hsize}{!}{$EI(\boldsymbol{p}) = 
    \begin{cases}
    (\boldsymbol{\mu}_{\bar{\boldsymbol{q}}}(\boldsymbol{p}) - q^+ - \xi)\boldsymbol{\Phi}(\boldsymbol{Z}) +\boldsymbol{\sigma}_{\bar{\boldsymbol{q}}}^2(\boldsymbol{p})\boldsymbol{\phi}(\boldsymbol{Z}) & \text{if } \boldsymbol{\sigma}_{\bar{\boldsymbol{q}}}^2(\boldsymbol{p}) > 0\\
    0              & \text{if } \boldsymbol{\sigma}_{\bar{\boldsymbol{q}}}^2(\boldsymbol{p}) = 0
    \end{cases}$}
\end{equation}
where $\boldsymbol{Z} = \frac{\boldsymbol{\mu}_{\bar{\boldsymbol{q}}}(\boldsymbol{p}) - q^+}{\boldsymbol{\sigma}_{\bar{\boldsymbol{q}}}^2(p)}$ if $\boldsymbol{\sigma}_{\bar{\boldsymbol{q}}}^2(\boldsymbol{p}) > 0$ else $0$; $\boldsymbol{\Phi}$ and $\boldsymbol{\phi}$ are the Probability Density Function (PDF) and Cumulative Density Function (CDF) of standard normal distribution, respectively and $q^+ = {q}^+(\boldsymbol{\mathcal{I}}(\boldsymbol{p}))$ is the best observed image quality so far. The exploration and exploitation during optimization is balanced by parameter $\xi$, and high $\xi$ means more exploration or less exploitation.
\subsection{Image quality estimation:} \label{sec:quality_estimator}
\begin{figure}[!ht]
	\centering
	\includegraphics[trim=0cm 8cm 11.5cm 0cm,clip,width=\linewidth]{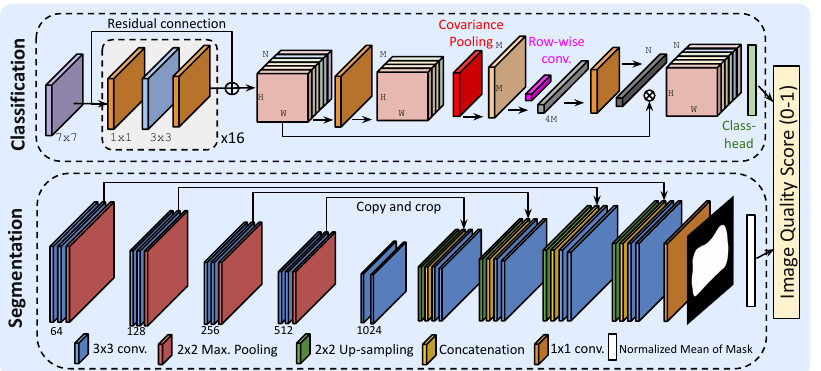}
	\caption{D-CNNs for image quality estimators}
	\label{fig:iq_estimate}
\end{figure}

We proposed two image quality estimators for estimating real-time diagnostic ultrasound image quality in A-RUS, which are based on: $(1)$ Classification and $(2)$ Segmentation of ultrasound image, as shown in Fig. \ref{fig:iq_estimate}.
\vspace{-1em}
\vspace{-1em}
\subsubsection{Image classification}
We propose using a Deep Convolution Neural Network (D-CNN) based classifier to extract the ultrasound image features and classify them based on the diagnostic quality. Ultrasound image quality assessment requires rich feature extraction for classifying the images that look quite similar but differ a lot in terms of image quality. In recent work, Song \textit{et al.} \cite{song2022medical} proposed a Bilinear Convolutional Neural Network (B-CNN) for fine-grained classification of breast ultrasound image quality. We propose a statistical enhancement to this work to improve its performance for ultrasound images having complex anatomical structures. Recently, Li \textit{et al.} \cite{gao2019global} proposed Second-order Pooling (SoP) block for enhancing the non-linear modeling ability of D-CNNs. Inspired by their work, we perform the SoP along spatial dimension at the end of the D-CNN network. First, we reduce the number of channels of feature volume $\mathcal{\boldsymbol{X}} \in \mathbb{R}^{H \times W \times N}$ from $N$ to $M$ using $1 \times 1$ convolution. Then, $\mathcal{\boldsymbol{X}}$ is reshaped to $\boldsymbol{X} \in \mathbb{R}^{M \times N}$ where $M = H \times W$. Later, we compute the covariance matrix $C_{M \times M} = (\boldsymbol{X} - \bar{\boldsymbol{X}})(\boldsymbol{X} - \bar{\boldsymbol{X}})^T$, reshaped it to $1 \times M \times M$ and passed it through convolution layer with $4M$ kernels of size $1 \times M$. The resulting tensor $1 \times 1 \times 4M$ is then resized to $1 \times 1 \times M$ tensor, $\boldsymbol{w}_M$, using a $1 \times 1$ convolution. Finally, weight vector $\boldsymbol{w}_M$ is multiplied with $\boldsymbol{X}$ to obtain the resultant feature volume $\boldsymbol{V}_M \in \mathbb{R}^{H \times W \times N}$. The base network used in proposed D-CNN is Residual Network (ResNet50). We trained the network using the Categorical Cross Entropy as a loss function.

\subsubsection{Image segmentation}
To evaluate the image quality based on image segmentation, we used U-net \cite{ronneberger2015u}, which is a state-of-the-art segmentation network for biomedical image segmentation. The network consists of two 2D convolution layer (Conv2D), followed by a Batch Normalization layer, Rectified Linear Unit (ReLU) as activation function, a max. pooling layer (i.e down sampling) of $2 \times 2$ in the contracting path. Each Conv2D layer is padded convolution with kernel size as $3 \times 3$. Similarly, for the expansive path, an up-sampling layer of $2 \times 2$ was added after each Conv2D layer. The number of filters along the contracting path doubled with each successive convolutional layer. The inverse of this pattern was replicated in the expanding path, resulting in the same number of filters in the last $3\times3$ convolutional layer of expanding path. In our dataset images, the bladder boundaries are blurred and irregular, making its segmentation rather challenging. Hence, we have used multiple loss functions to train the network, so that the model would be enforced to learn fine-grained features of the bladder \cite{raina2023slim}. We used the combinations of three loss functions, Dice coefficient (DC) and Jaccard Index (JI) and Binary Cross-Entropy (BCE), formulated as $\boldsymbol{L}_{DJB} = \boldsymbol{L}_{DC} + \boldsymbol{L}_{JI} + \boldsymbol{L}_{BCE}$, where
\begin{equation*}
    \boldsymbol{L}_{DC} = 1-\left ( \frac{2 \boldsymbol{y}_{t} \boldsymbol{y}_{p} + \boldsymbol{s}} {\boldsymbol{y}_{t} + \boldsymbol{y}_{p} + \boldsymbol{s}} \right), \boldsymbol{L}_{JI} = \boldsymbol{1} -  \frac{\boldsymbol{y}_{t} \boldsymbol{y}_{p} + \boldsymbol{s}} {\boldsymbol{y}_{t} + \boldsymbol{y}_{p} - \boldsymbol{y}_{t} \boldsymbol{y}_{p} + \boldsymbol{s}}
\end{equation*}
\begin{equation*}
    \boldsymbol{L}_{BCE} = -\boldsymbol{y}_{t} log(\boldsymbol{y}_{p})-(1-\boldsymbol{y}_{t}) log(1 - \boldsymbol{y}_{p})
\end{equation*}
where $\boldsymbol{s}$, $\boldsymbol{y}_{t}$ and $\boldsymbol{y}_{p}$ represent smoothness constant, ground truth and predicted mask, respectively. The model will finally output the predicted mask of the given input ultrasound image. We take normalized mean of the predicted mask as the quality score, the value of which lies between $0$ and $1$.
\subsubsection{Dataset} 
We collected ultrasound image dataset from Urinary Bladder (UB) phantom (YourDesignMedical, USA). A total of $2290$ phantom images were collected. All images have been resized to $224 \times 224$ shape. The training and validation set consists of $2061$ and $229$ images, respectively. For classification, the ground truth quality of images is an integer score between $1-5$, based on an internationally prescribed generalized $5$-level absolute assessment scale \cite{duan20215g}. The training and validation set contains an equal distribution of images for each quality type. A score of $1$ means no appearance of the UB with unacceptable artifacts and $5$ means that the clear depiction of the UB with distinct boundaries and acceptable artifacts, depicting a high diagnostic accuracy. A poor-quality image either contains noise or motion artifacts, blurred images, indistinct boundaries, obscuring the posterior or anterior sections of the UB. Later, we normalized the quality score in the range $0-1$ for comparison with segmentation-based quality score, where $0$ denotes a score when the force value of probe along z-axis is below minimum required for appropriate contact. For segmentation, we annotated using SuperAnnotate (https://superannotate.com/). A single polygon with multiple points is drawn in every image, spanning the shape of UB. These polygons were then used to generate corresponding ground truth masks. 

\subsection{Robotic controller}
The robotic controller will move the probe to the new pose $\boldsymbol{p} = [x, y, z, roll, pitch, yaw]$ given by BO. The hybrid position-force control is used for controlling the robot, where only $z-$axis is under force control, denoted as $f_z$. For the safety of phantoms, the force limits have been set to $20 N$. 
\section{Results and Discussions} \label{sec:results}
\subsection{Experimental setup} \label{sec:exp_setup}
We validate our framework on an experimental setup of robotic ultrasound system, consisting of a $7$-DoF Sawyer collaborative robotic arm (Rethink Robotics, Germany) with a Micro Convex MC10-5R10S-3 probe attached to its end-effector. The probe is connected to the Telemed Ultrasound machine (Telemed Medical Systems, Italy), which captures the ultrasound image. The machine is connected to the laptop for displaying and transmitting the image for processing in BO. The ultrasound scanning is conducted on a urinary bladder phantom. 
\begin{figure}[!ht]
	\centering
	\includegraphics[trim=0cm 4.4cm 7cm 0cm,clip,width=0.9\linewidth]{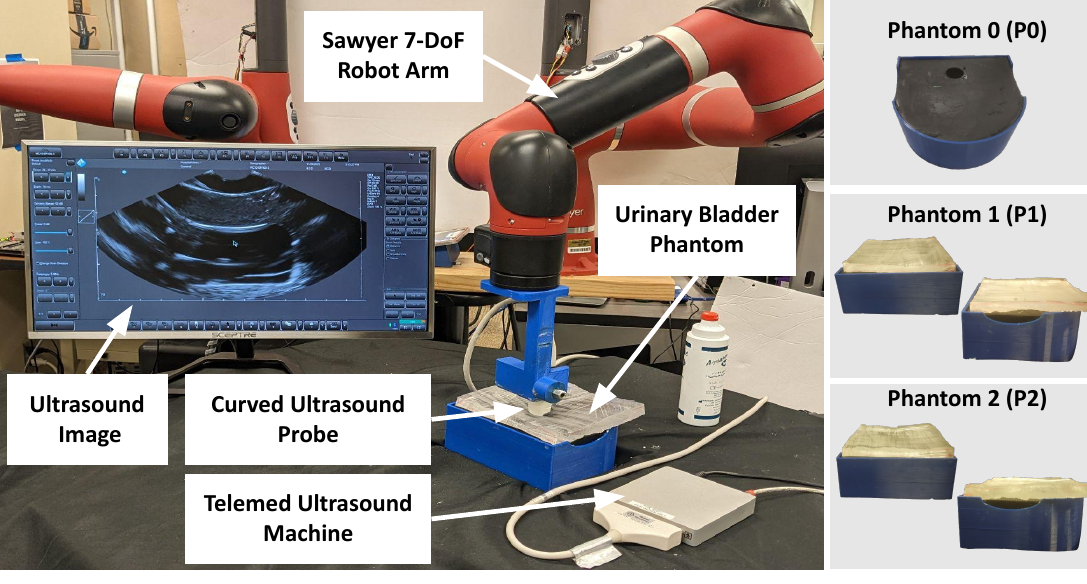}
	\caption{Experimental setup of robotic ultrasound system with three different phantoms of the urinary bladder.}
	\label{fig:exp_setup}
\end{figure}
In order to validate the proposed framework for inter-patient ultrasound scanning, we modified the phantom $P0$ using the layers of ballistic gel. We used $0.38$ inch layer for $P1$ and $0.63$ inch layer for $P2$, as shown in Fig. \ref{fig:exp_setup}. The BO has been implemented in  Python $3.8$. The DK neural network and IQEs explained in Section \ref{sec:methodology} has been trained using PyTorch $1.11$. For BO, we used $\xi=0.1$ and region limits $A$ as $x \in (-0.05,0.05)m$, $ y \in (-0.02,0.02)m$, $f_z \in (5-20)N$, $roll \in (-0.2,0.2)rad$, $pitch \in (-0.2,0.2)rad$, and $yaw \in (-0.5,0.5)rad$. The orientation limits have been decided to ensure collision-free scanning of phantoms.
\subsection{Analyzing BO performance with Deep kernel}
In order to analyze the effectiveness of the proposed framework, we compared the performance of BO that used our proposed neural network based Deep kernel with a standard Radial Basis Function (RBF) kernel. We have analysed our results for two feedback functions, as described in Section \ref{sec:quality_estimator}: (i) Image quality score obtained from classification network $(q_c)$, which is a non-smooth feedback; (ii) Image segmentation score obtained from segmentation network $(q_s)$, which is a smooth feedback function. 
\begin{figure}[b]
\centering
\subfloat[]{\includegraphics[trim=0cm 0.8cm 0cm 0.6cm,clip, width=0.5\linewidth]{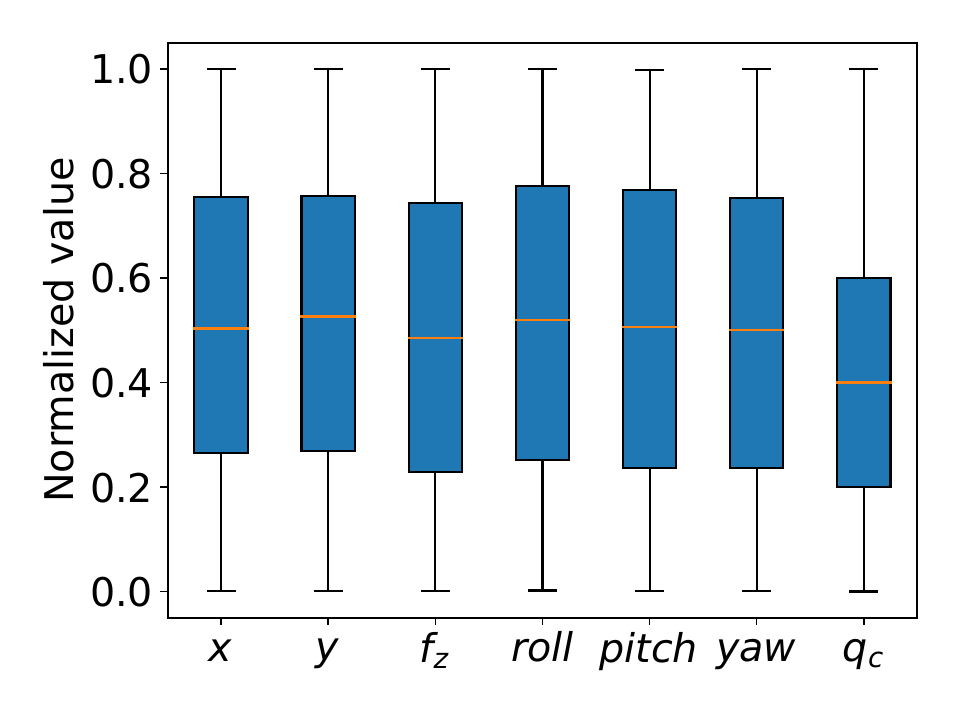}%
\label{fig:dk_data}}
\subfloat[]{\includegraphics[trim=0cm 0.8cm 0cm 0.6cm,clip,width=0.5\linewidth]{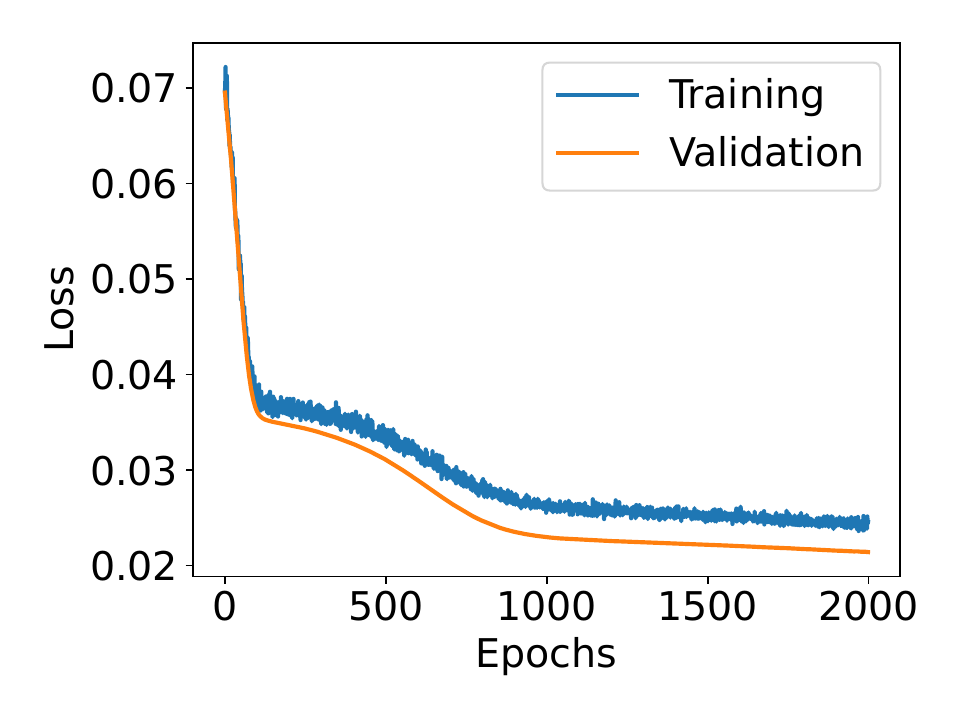}%
\label{fig:learning_curve}}
\caption{(a) Distribution of probe poses sampled for training; (b) learning curve of deep kernel's neural network}
\label{fig:dk_related}
\end{figure}
The dataset for deep kernel has been sampled from phantoms $P0$ and $P1$, while no dataset has been taken from $P2$, which is our test-case phantom. 
For training the deep kernel, we sampled $1200$ probe poses, $600$ for each of the phantom $P0$ and $P1$, using latin hypercube sampling and scanned the phantoms using sampled probe poses. 
The distribution of probe-poses sampled for each pose variable and corresponding quality is shown in Fig. \ref{fig:dk_data}. The distribution shows that probe control parameters have been widely explored. However, out of the $1200$ sampled poses, only $110$ resulted in a quality greater than $0.8$. This means that random sampling has only $9.2\%$ chance of locating a high image quality region. This shows that acquiring high-quality images requires all sets of probe parameters to be accurate, which highlights the complexity of this optimization problem. The learning curve for the deep kernel's neural network is shown in Fig. \ref{fig:learning_curve}. 
\begin{figure}[!ht]
\centering
\subfloat[]{\includegraphics[trim=0cm 0.8cm 0.5cm 0.8cm,clip,width=0.5\linewidth]{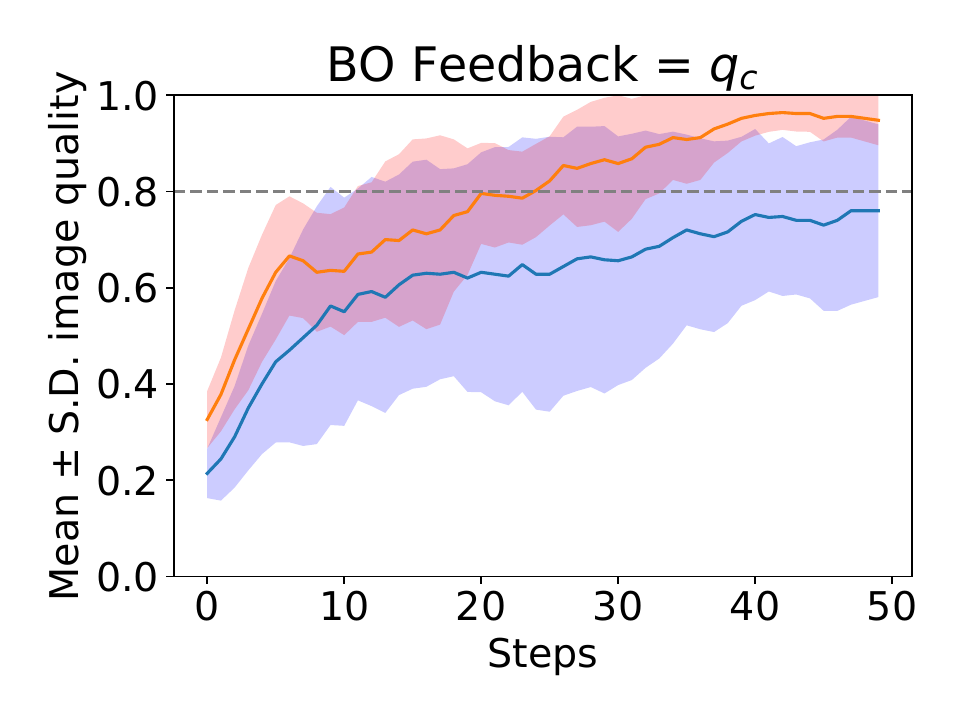}%
\label{fig:p1_class}}
\subfloat[]{\includegraphics[trim=0cm 0.8cm 0.5cm 0.8cm,clip, width=0.5\linewidth]{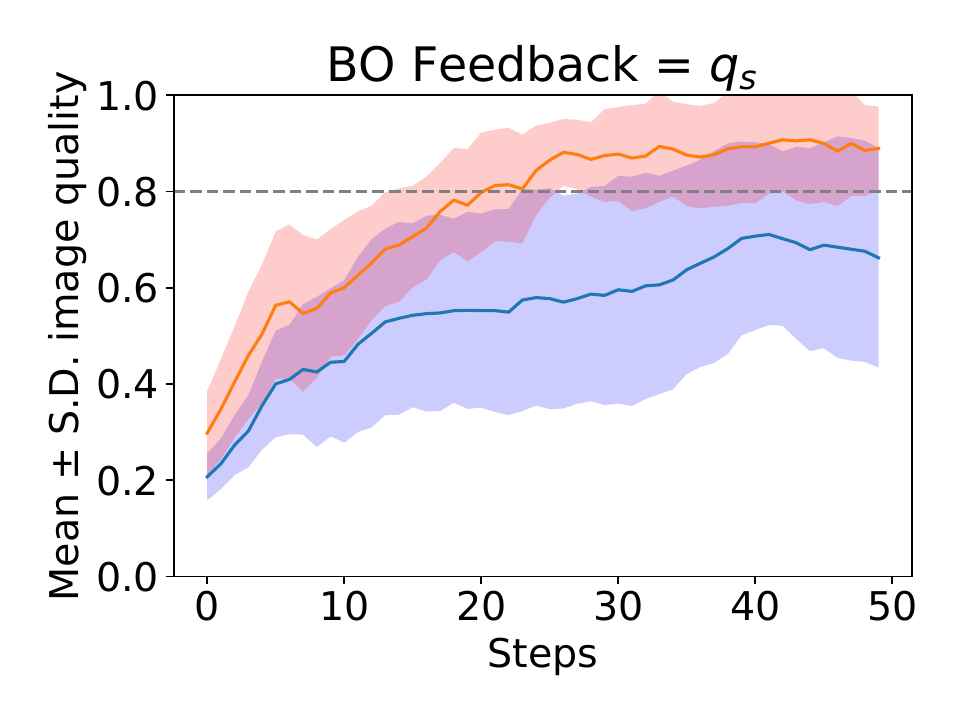}%
\label{fig:p1_seg}}\\
\subfloat[]{\includegraphics[trim=0cm 0.8cm 0.5cm 0.6cm,clip,width=0.5\linewidth]{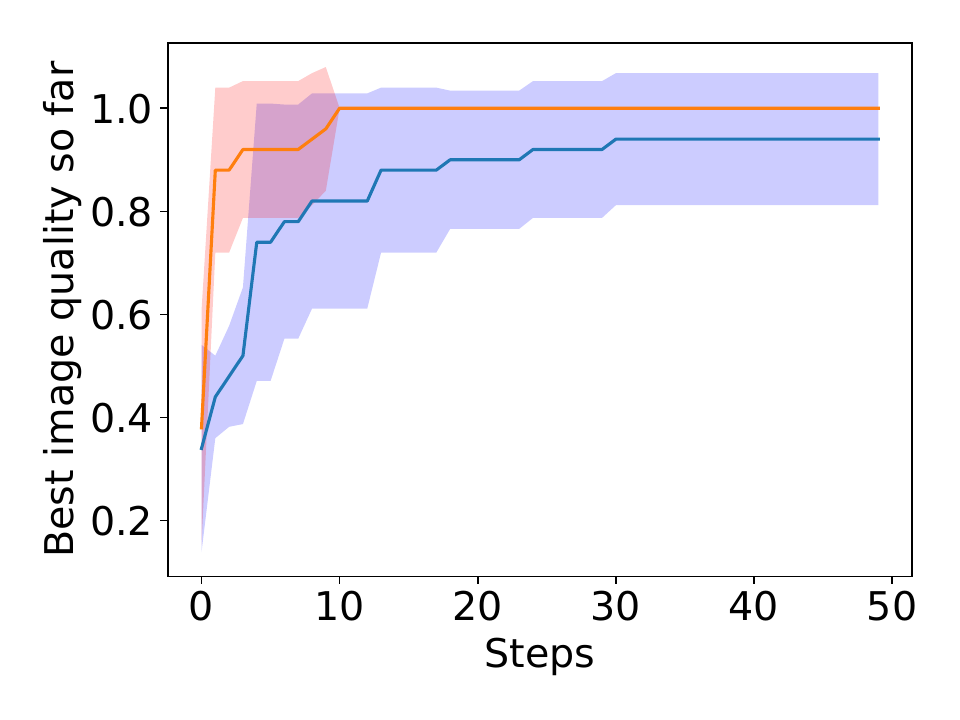}%
\label{fig:p1_class_best}}
\subfloat[]{\includegraphics[trim=0cm 0.8cm 0.5cm 0.6cm,clip,width=0.5\linewidth]{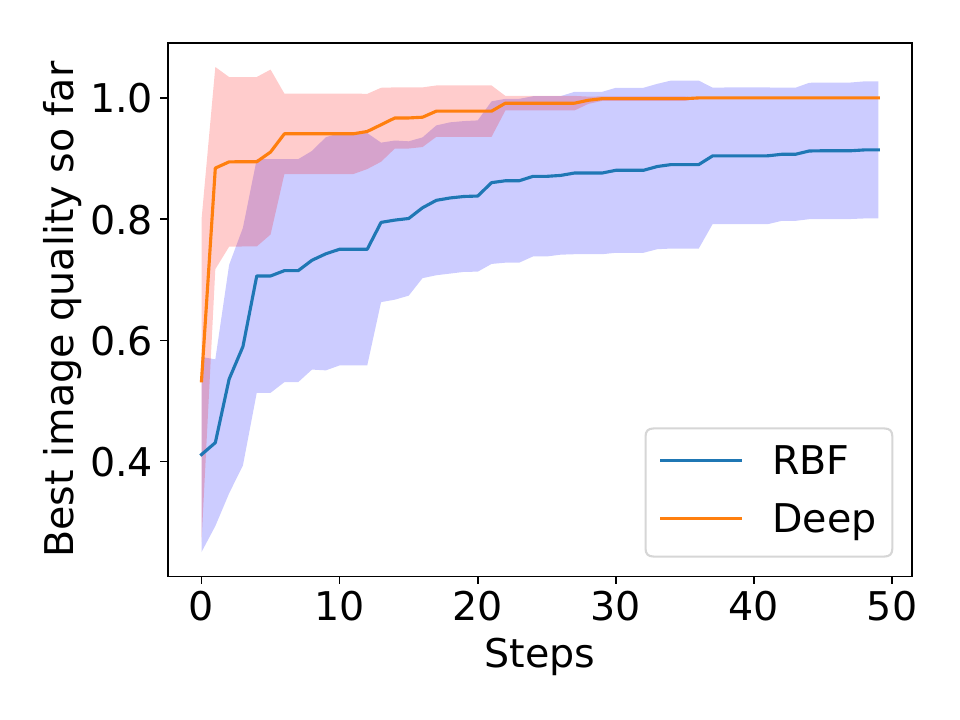}%
\label{fig:p1_seg_best}}\\
\subfloat[]{\includegraphics[trim=0cm 0.8cm 0.5cm 0.6cm,clip,width=0.5\linewidth]{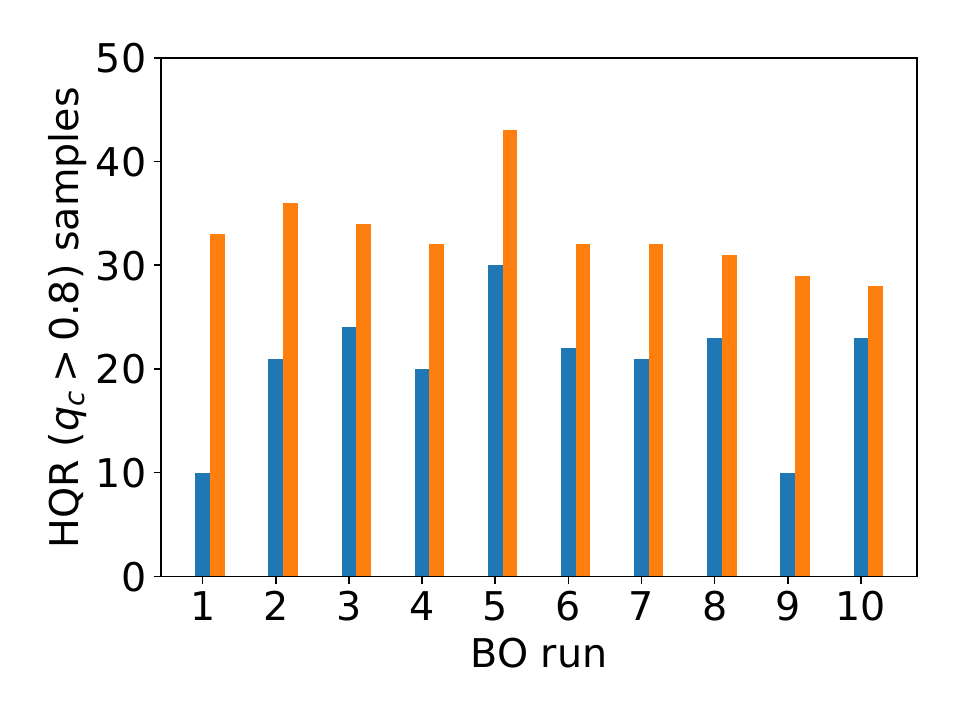}%
\label{fig:p0p1_class_best_freq}}
\subfloat[]{\includegraphics[trim=0cm 0.8cm 0.5cm 0.6cm,clip,width=0.5\linewidth]{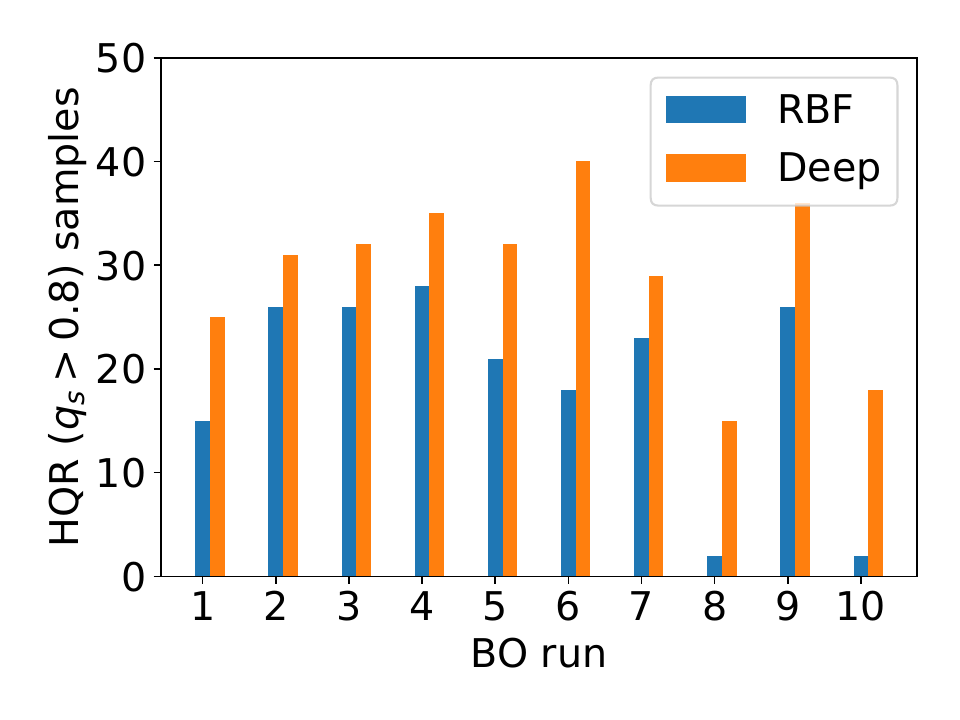}%
\label{fig:p0p1_seg_best_freq}}
\caption{Analysis of BO runs on phantom $P0$ and $P1$ using classification $q_c$ (a,c,e)  and segmentation $q_s$ (b,d,f) feedback. Line show mean and shaded region shows $\pm$S.D over $10$ runs}
\label{fig:bo_pop1}
\end{figure}

In total, we conducted $60$ runs of BO, $5$ each for $3$ phantoms and $2$ feedback functions. Each run of BO has $50$ steps, which is the termination condition. In this section, we denote the quality score greater than $0.8$ as \textit{high quality region (HQR)}.
The average image quality achieved during the runs of BO for phantom $P0$, $P1$ are shown in Fig \ref{fig:p1_class} and \ref{fig:p1_seg}. It indicates that BO with RBF kernel maintains a low value of average image quality compared to deep kernel over all steps of BO, and hence it could not locate the HQR reliably even after $50$ steps. Further, the large values of standard deviation (represented by the shaded region on both sides of the mean line) for RBF kernel show its extensive exploratory nature. In contrast, deep kernel maintains high mean quality with low standard deviations, which shows its better sample efficiency. The deep kernel also succeeded in locating the HQR in first $5-10$ steps in all runs of BO, as shown in Fig. \ref{fig:p1_class_best} and \ref{fig:p1_seg_best}. Further, Fig. \ref{fig:p0p1_class_best_freq} and \ref{fig:p0p1_seg_best_freq} shows that deep kernel has $61.76\%$ and $56.68\%$ more sample in HQR than RBF for $q_c$ and $q_s$ feedback, respectively, in 10 runs of BO. 
\begin{figure}[t]
\centering
\subfloat[]{\includegraphics[trim=0cm 0.8cm 0.5cm 0.7cm,clip,width=0.5\linewidth]{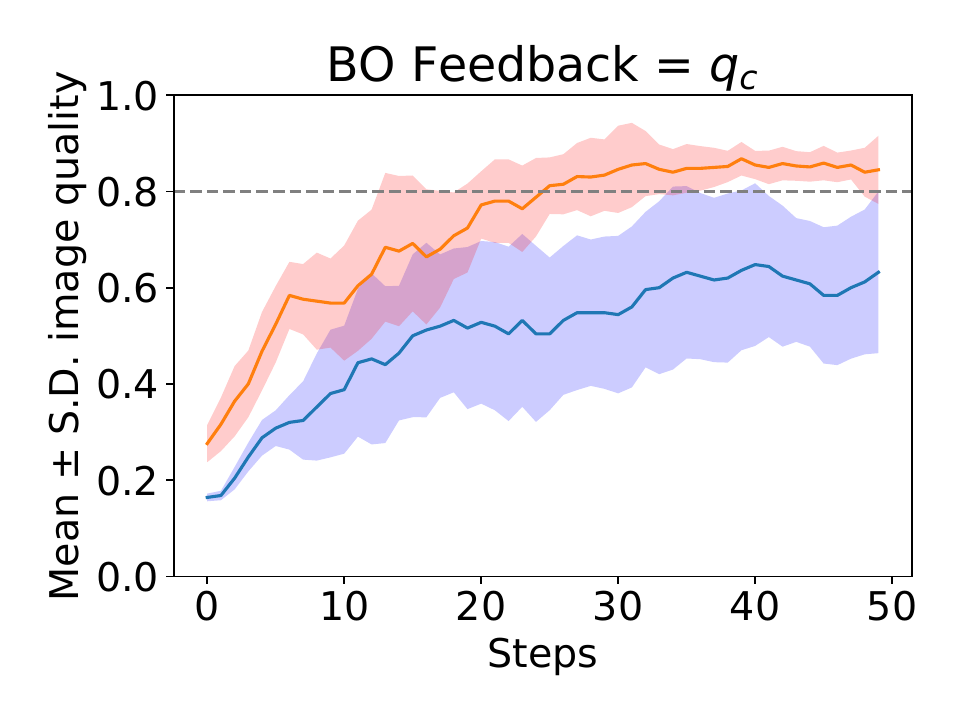}%
\label{fig:p2_class}}
\subfloat[]{\includegraphics[trim=0cm 0.8cm 0.5cm 0.7cm,clip, width=0.5\linewidth]{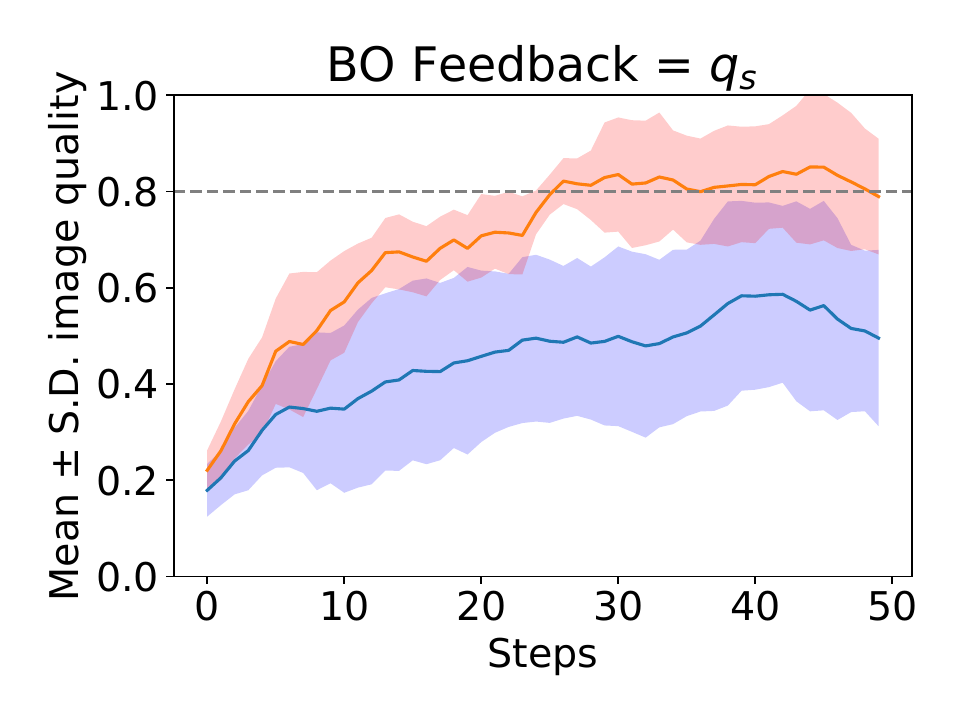}%
\label{fig:p2_seg}}\\
\subfloat[]{\includegraphics[trim=0cm 0.8cm 0.5cm 0.6cm,clip,width=0.5\linewidth]{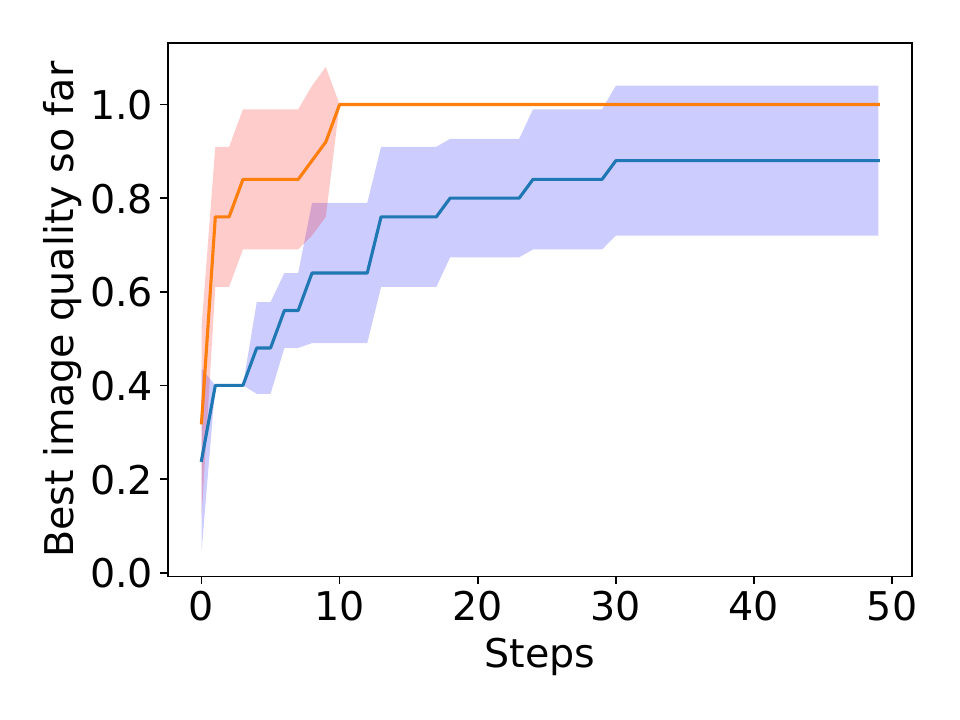}%
\label{fig:p2_class_best}}
\subfloat[]{\includegraphics[trim=0cm 0.8cm 0.5cm 0.6cm,clip,width=0.5\linewidth]{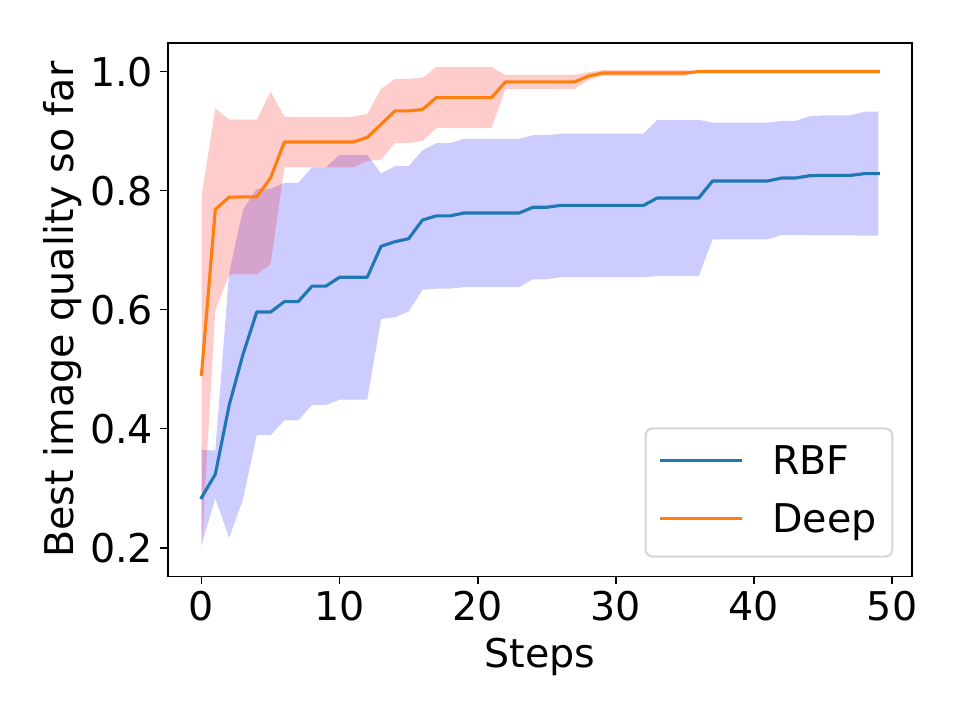}%
\label{fig:p2_seg_best}}\\
\subfloat[]{\includegraphics[trim=0cm 0.8cm 0.5cm 0.6cm,clip,width=0.5\linewidth]{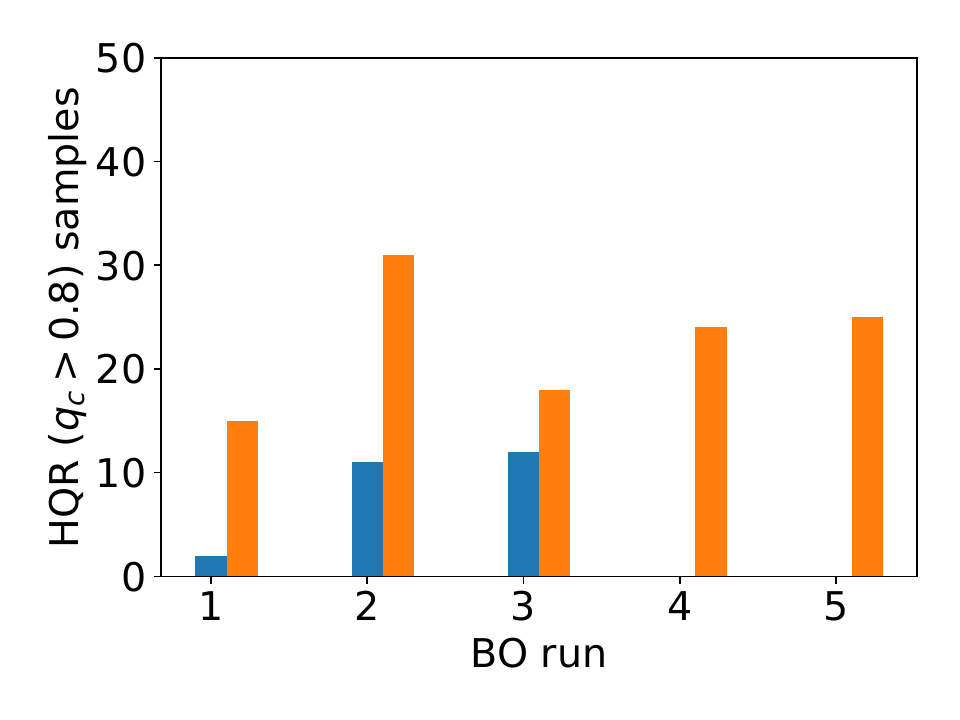}%
\label{fig:p2_class_best_freq}}
\subfloat[]{\includegraphics[trim=0cm 0.8cm 0.5cm 0.6cm,clip,width=0.5\linewidth]{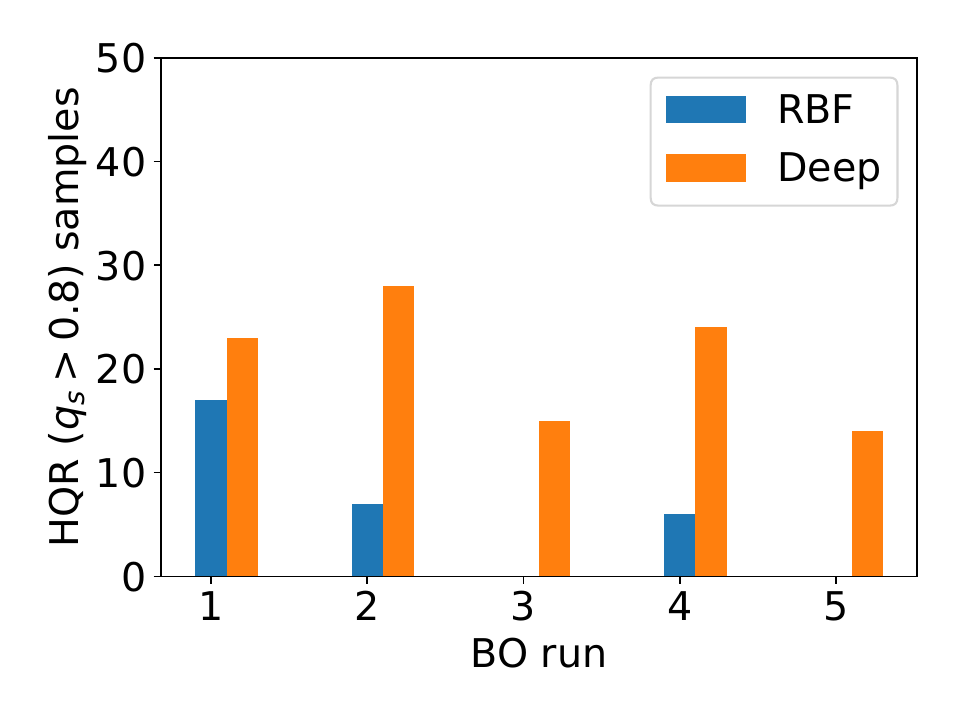}%
\label{fig:p2_seg_best_freq}}
\caption{Analysis of BO runs on phantom $P2$ using classification $q_c$ (a,c,e) and segmentation $q_s$ (b,d,f)  as feedback. Line show mean and shaded region shows $\pm$S.D over $10$ runs}
\label{fig:bo_p2}
\end{figure}

For phantom $P2$, the average values in Fig. \ref{fig:p2_class} and \ref{fig:p2_seg} show that BO with RBF kernel could not locate the HQR reliably (high S.D.). However, BO with deep kernel reached the HQR in approximately $10-30$ steps, as shown in Fig. \ref{fig:p2_class_best} and \ref{fig:p2_seg_best}. In contrast, the RBF kernel took more than $30$ steps to locate the HQR reliably. Moreover, out of the $5$ runs, RBF kernel could only locate the HQR in $3/5$ runs for $q_c$ and $q_s$, as shown in Fig. \ref{fig:p2_class_best_freq} and Fig. \ref{fig:p2_seg_best_freq}. Further, RBF has less number of HQR samples in comparison to deep kernel across all runs of BO. These results indicate that the deep kernel improves the sample efficiency of BO and its performance is independent of its training dataset. 
\subsection{Analysis of image quality estimators}
We evaluated the classification network on a $10$-fold cross-validation for classifying the urinary bladder phantom images based on their quality. To avoid over-fitting on the small dataset, we applied random horizontal flipping and normalization data augmentations. The values of precision, recall, and accuracy for five classes of quality are shown in Table \ref{tab:usqnet_evaluate}. We compared the proposed network, ResNet50 with Spatial SoP (RN50+SSoP), with the ResNet50 with Bilinear Pooling (RN50+BP) \cite{song2022medical}, to validate the performance enhancement. Results show that the RN50+SSoP has an average accuracy of $98.2\pm0.5$, which is $1.42\%$ more than the average accuracy of RN50+BP. The class-wise analysis suggest that RN50+SSoP has significant gains over the RN50+BP, achieving $1.46\%$, $2.06\%$, $2.95\%$ and $0.91\%$ more accuracy for quality scores $2$, $3$, $4$ and $5$, respectively. Hence, it is validated that the proposed network enhanced the accuracy of ultrasound image quality classification.
\renewcommand{\arraystretch}{1.05}
\begin{table}[!ht]
  \centering
    \caption{Comparison of the classification-based IQE with the state-of-the art network on 10-fold cross validation}
  \resizebox{\linewidth}{!}{\begin{tabular}{ccccccc}
    \toprule
    \multirow{2}{*}{\textbf{Score}}  & \multicolumn{3}{c}{\textbf{RN50+BP} \cite{song2022medical}} &  \multicolumn{3}{c}{\textbf{RN50+SSoP (Proposed)}} \\
    \cmidrule(lr){2-4}\cmidrule(lr){5-7}
    ~ & \textbf{Precision} & \textbf{Recall} & \textbf{Accuracy} & \textbf{Precision} & \textbf{Recall} & \textbf{Accuracy} \\
    \midrule
    $\boldsymbol{1}$ & $99.7\pm0.3$ & $99.4\pm0.7$ & $99.5\pm0.4$ &  $99.7\pm0.5$ &  $99.9\pm0.3$ & $99.8\pm 0.4$ \\
    $\boldsymbol{2}$ & $93.9\pm2.1$ & $95.8\pm1.7$ & $94.8\pm1.3$ &  $95.5 \pm 1.7$ &  $97.0 \pm 1.5$ & $96.2 \pm 1.3$ \\
    $\boldsymbol{3}$ & $94.6\pm1.3$ & $95.5\pm1.9$ & $95.0\pm1.1$ &  $97.8 \pm 1.3$ &  $96.3 \pm 1.6$ & $97.0 \pm 1.2$ \\
    $\boldsymbol{4}$ & $96.3\pm1.4$ & $94.8\pm1.6$ & $95.5\pm0.7$ &  $98.7 \pm 0.9$ &  $98.1 \pm 0.4$ & $98.4 \pm 0.3$ \\
    $\boldsymbol{5}$ & $97.3\pm0.8$ & $98.9\pm0.3$ & $98.0\pm0.1$ &  $98.7 \pm 0.6$ &  $99.1 \pm 0.2$ & $98.9 \pm 0.1$ \\
    \midrule
    \textbf{Avg.} & $\boldsymbol{96.5\pm0.5}$ & $\boldsymbol{97.1\pm0.5}$ & $\boldsymbol{96.8\pm0.5}$ &  $\boldsymbol{98.2} \pm \boldsymbol{0.5}$ &  $\boldsymbol{98.2} \pm \boldsymbol{0.5}$ & $\boldsymbol{98.2} \pm \boldsymbol{0.5}$ \\

    \bottomrule
  \end{tabular}}
  \label{tab:usqnet_evaluate}
\end{table}

\begin{figure}
		\centering
		\includegraphics[trim=0cm 9.7cm 3cm 0.5cm,clip,width=\linewidth]{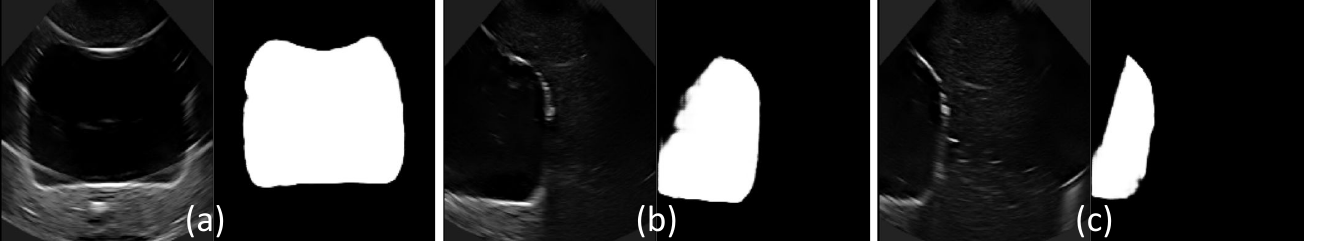}  
		\caption{The predicted masks when bladder appears (a) at full size, (b) approx. half of its size, and (c) one-fourth of its size} 
  \label{fig:seg_results}
\end{figure}

For the segmentation network, the qualitative results are shown in Fig. \ref{fig:seg_results}. We further evaluated this network quantitatively using the standard evaluation matrices, including Precision (P), Recall (R), F1 score (F1), Dice Coefficient (DC), and Intersection over Union (IoU). The values of these evaluation metrics for the 10-fold cross-validation and comparison with standard dice loss function ($\boldsymbol{L}_{DC}$) are given in Table \ref{tab:seg_res}. The proposed loss function, $\boldsymbol{L}_{DJB}$ outperforms $\boldsymbol{L}_{DC}$ on P, R, F1, DC and IoU by $3.2\%$, $4.7\%$, $2.6\%$, $2.5\%$ and $4.8\%$, respectively.  Hence, the U-net guided with the proposed multi-loss function is an effective method for the segmentation of complex structures in the ultrasound image. 
\vspace{-3mm}
\begin{table}[!ht]
\caption{Comparison of the segmentation-based IQE for standard dice loss and proposed multi-loss function}
  \label{tab:seg_res}
  \resizebox{\linewidth}{!}{\begin{tabular}{ccccccccccc}
    \toprule
    \textbf{Loss} & \textbf{P} & \textbf{R} & \textbf{F1} & \textbf{DC} & \textbf{IoU} \\
    \midrule
    $\boldsymbol{L}_{DC}$ & $95.6$$\pm6.09$ & $92.7$$\pm5.55$ & $93.9$$\pm2.99$ & $93.7$$\pm3.04$ & $88.4$$\pm5.41$\\   
    $\boldsymbol{L}_{DJB}$ & $98.8$$\pm0.26$ & $97.3$$\pm2.02$ & $96.4$$\pm1.57$ & $96.1$$\pm2.10$ & $92.9$$\pm1.81$ \\
    \bottomrule
\end{tabular}}
\end{table}
\vspace{-2mm}
\section{Conclusion}
We proposed a sample-efficient Bayesian Optimization (BO) framework using deep kernel and image quality estimators for optimizing the 6-dimensional robotic ultrasound controller. The deep kernel of Gaussian process model learns a one-dimensional embedding representing image quality for a given probe pose. The two image quality estimators (IQEs) based on classification and segmentation utilizing a deep convolutional neural network are proposed for real-time quality assessment in RUS. We validated the framework on three urinary bladder phantoms and demonstrated over $50\%$ increase in sample efficiency over the standard kernel for both IQEs as feedback to BO. In future work, we would like to validate it for complicated surface phantoms and \textit{in-vivo} trials for a clinical procedure \cite{raina2021comprehensive,chandrashekhara2022robotic}. We will explore approaches for improving the IQEs and learning the kernel from multiple expert demonstrations. We would also investigate integrating the deep kernel with a pre-trained prior quality model similar to \cite{zhu2022automated} for further improvement in the sample efficiency of BO. 
\bibliography{references} 
\bibliographystyle{ieeetr}
\end{document}